\documentclass{esannV2}
\usepackage{xcolor}
\usepackage{graphicx}
\usepackage[latin1]{inputenc}
\usepackage{amssymb,amsmath,array}
\usepackage{subcaption}
\usepackage{url}

\usepackage{booktabs}
\usepackage{adjustbox}

\begin{document}
\title{Robust Evolutionary Multi-Objective Network Architecture Search for Reinforcement Learning (EMNAS-RL)}

\author{Nihal Acharya Adde$^1$, Alexandra Gianzina$^1$, Hanno Gottschalk$^2$, Andreas Ebert$^1$
%
%
\vspace{.3cm}\\
%
1- Volkswagen Group Innovation, Volkswagen AG, Wolfsburg, Germany 
%
\vspace{.1cm}\\
2- Institute of Mathematics, TU Berlin, Germany\\
}

\maketitle

\begin{abstract}
This paper introduces Evolutionary Multi-Objective Network Architecture Search (EMNAS) for the first time to optimize neural network architectures in large-scale Reinforcement Learning (RL) for Autonomous Driving (AD). EMNAS uses genetic algorithms to automate network design, tailored to enhance rewards and reduce model size without compromising performance. Additionally, parallelization techniques are employed to accelerate the search, and teacher-student methodologies are implemented to ensure scalable optimization. This research underscores the potential of transfer learning as a robust framework for optimizing performance across iterative learning processes by effectively leveraging knowledge from earlier generations to enhance learning efficiency and stability in subsequent generations. Experimental results demonstrate that tailored EMNAS outperforms manually designed models, achieving higher rewards with fewer parameters. The findings of these strategies contribute positively to EMNAS for RL in autonomous driving, advancing the field toward better-performing networks suitable for real-world scenarios.
\end{abstract}

\section{Introduction}
Neural Architecture Search (NAS) automates the traditionally manual and time-intensive process of designing neural network architectures. NAS algorithms use different methods like Reinforcement Learning (RL), Evolutionary Algorithms (EA), and Gradient-based Optimization (GO), to explore potential architectures to identify the most effective designs for specific tasks. EAs are particularly effective, consistently achieving state-of-the-art performance \cite{real2019regularized}, with superior anytime performance \cite{real2019regularized} that enables the discovery of smaller models compared to RL and addresses multi-objective problems in NAS \cite{termritthikun2021eeea,lu2019nsga}. This approach seeks high-performing models, reflected in RL rewards, while minimizing constraints such as parameter size for memory efficiency and Floating Point Operations Per Second (FLOPS) to meet power consumption or latency requirements.

This paper explores the use of EAs to optimize network architectures for an RL task in Autonomous Driving (AD), aiming to outperform human-designed baselines. A teacher-student framework is employed to transfer learned knowledge across generations, where student networks are trained via behavior cloning \cite{inproceedings} using the top-performing teacher network from the previous generation. 

\subsection{Autonomous Driving Task}
\begin{figure}[h]
    \centering
    \includegraphics[width=\linewidth]{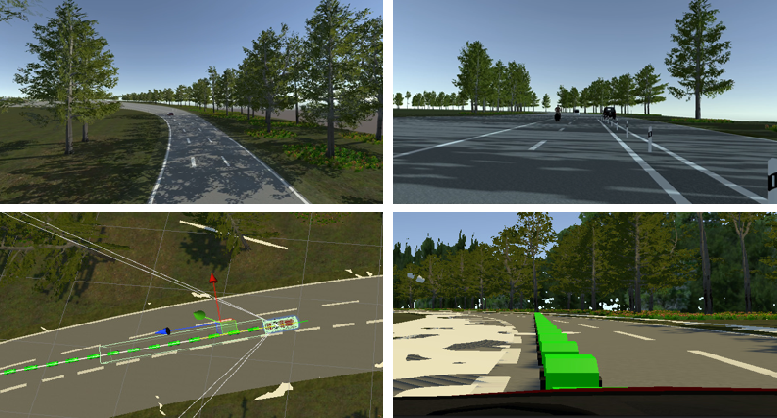}
    \caption{Snapshots from different angles of the Unity simulator showcasing the car driving through the oval track. Green boxes represent trajectories, which are sequentially arranged by the RL algorithm to ensure optimal driving behavior.}
    \label{fig:driving_snippets}
\end{figure}

The AD task is trained in the Unity 3D simulator \cite{Unity} using the Proximal Policy Optimization (PPO) algorithm \cite{schulman2017proximal}. In this setup, multiple agents navigate a three-lane oval track, aiming for smooth, collision-free driving with minimal abrupt movements. Agents use semantic segmentation from a forward-facing camera to predict trajectory points, which are processed by an external vehicle controller for maneuvering. Reward structures incentivize precise trajectory placement and smooth, collision-free navigation. This study focuses on optimizing convolutional network architectures for this RL task. Figure \ref{fig:driving_snippets} illustrates four snapshots of the driving environment from various angles, showcasing scenarios encountered during training.

\subsection{General framework of Evolutionary Network Architecture Search}
\begin{figure}[!ht]
\centering
    \includegraphics[ width = \linewidth]{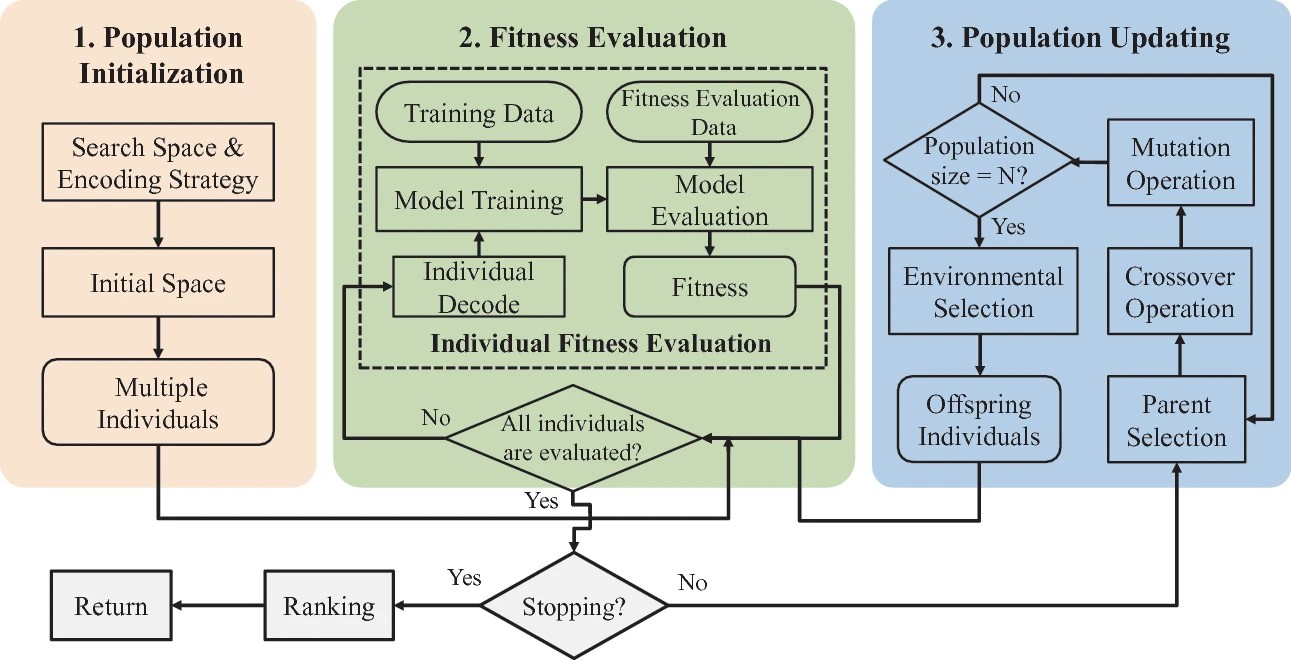}
 \caption{An illustration of the ENAS algorithm highlighting its three main phases: population initialization (light orange), fitness evaluation (light green), and population updating (light blue). .}
\label{fig:NAS}
\end{figure}

The general framework of Neural Architecture Search (NAS) using Evolutionary Algorithms (EAs) is inspired by natural evolution, employing population-based meta-heuristic optimization methods to address complex tasks and real-world challenges \cite{Lv2024}. EAs generate numerous solutions simultaneously and do not rely on gradient information or domain expertise, making them versatile and capable of finding global optimal solutions. These characteristics have made EAs increasingly popular to tackle the NAS problem, commonly referred to as Evolutionary NAS (ENAS). ENAS focuses on automating the design of deep neural networks, including architectures like Deep-CNNs \cite{9075201}. The process involves multiple iterations where architectures are generated, trained, and evaluated to guide subsequent iterations. A typical ENAS framework, as illustrated in Figure \ref{fig:NAS}, consists of three main phases: population initialization, fitness evaluation, and population updating \cite{Lv2024}. The iterative process concludes when a predefined stopping condition is met, marking the discovery of an optimal neural network architecture.
\begin{itemize}
    \item \textbf{Population Initialization:} In the first phase, a population of individuals is generated from a predetermined search space. Each individual is encoded to represent its genes (operation types and connections). Once initialized, the individuals proceed to evaluation.
    
    \item \textbf{Fitness Evaluation:} Each individual undergoes evaluation where its genes are decoded and trained. Fitness values are calculated, and individuals are ranked based on performance. If the stopping condition is satisfied, the process ends; otherwise, the algorithm moves to the next step.
    
    \item \textbf{Population Updating:} During this phase, parents are selected and genetic operators such as mutation and crossover are applied to generate offspring. The new generation of individuals is then sent back for evaluation, continuing the evolutionary process until the termination condition is met.
\end{itemize}

\section{Background}
Evolutionary Algorithms (EAs) are effective in NAS due to their ability to explore complex search spaces. Early methods like NeuroEvolution of Augmenting Topologies (NEAT) \cite{stanley2002NEAT} demonstrated the potential of genetic algorithms in evolving neural architectures. A comparison by \cite{real2019regularized} highlighted EA's superior performance over RL and random search, leading to models like AmoebaNet \cite{real2019regularized}. Recently, Multi-Objective Evolutionary Algorithms (MOEAs) have been applied to NAS, optimizing multiple objectives in parallel. Notably, NSGA-Net \cite{lu2019nsga} leverages MOEA to produce a Pareto front of architectures balancing accuracy with computational efficiency for image classification tasks. Early Exit Evolutionary Neural Architecture Search (EEEANet) \cite{termritthikun2021eeea} combines MOEA with early-exit population initialization, yielding compact architectures for resource-constrained devices. EEEANet uses tournament selection and non-dominated sorting with the NSGA-II algorithm \cite{deb2002fast}. While NAS has advanced broadly, its application to RL remains limited. Our research addresses this gap by applying EA-based NAS to optimize network architectures for RL in AD, aiming to improve efficiency and scalability.

\section{EMNAS-RL Methodology}
 




\begin{figure}[!ht]
\centering
    \includegraphics[height=100pt, width = \linewidth]{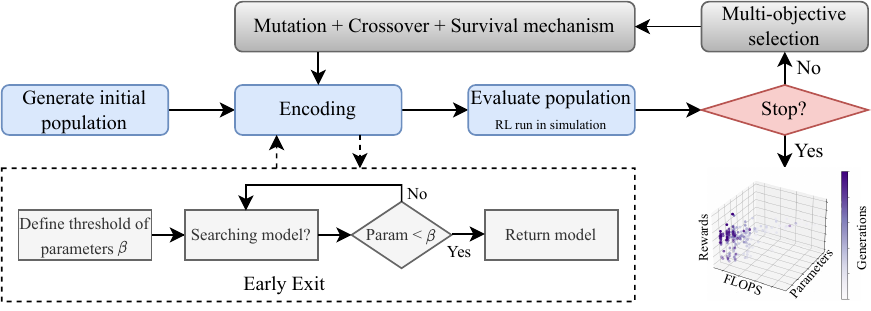}
 \caption{Multi-objective Evolutionary Algorithm with EEPI. Illustration of the early exit strategy, which ensures rejection of extremely large networks during initialization, favoring smaller and more efficient networks.}
\label{fig:EEEANet_algo_as_graphic}
\end{figure}

In this work, we build upon EEEANet \cite{termritthikun2021eeea}, using MOEA-based NAS as the foundation. EA, based on Genetic Algorithms (GA), iteratively evaluates a randomly initialized population of neural architectures using a fitness function to assess performance \cite{termritthikun2021eeea}. The population evolves over generations through crossover and mutation, where a population represents candidate architectures, and a generation is a single evolutionary iteration. Figure \ref{fig:EEEANet_algo_as_graphic} illustrates the algorithm framework.

In MOEA, the fitness function combines multiple objectives: reward (model effectiveness), computational cost (FLOPS), and model complexity (parameters), minimizing negative reward while balancing efficiency, as shown in Eq. \eqref{eq:1}. 
\begin{equation} \label{eq:1}
    f(x) = \min \; \{\textbf{-Reward}(x), \textbf{FLOPS}(x), \textbf{Params}(x)\}, x \in X 
\end{equation}
where $x$ is an individual architecture in the set $X$ of a given generation. These objectives are normalized and equally weighted to optimize for limited computational resources. The NSGA-II algorithm ranks the population based on these objectives, evolving a diverse set of non-dominated solutions guided by Pareto optimality principles \cite{van1998evolutionary}. We use a Pareto front-based tournament selection, dividing the population into subsets to compare fitness values, with winners reproducing through crossover and mutation to enhance diversity \cite{termritthikun2021eeea}. Crossover, mutation, and survival mechanisms drive the creation of new generations, retaining promising architectures while exploring the search space \cite{termritthikun2021eeea}. Mutation alters genetic information to explore new search space regions, while crossover combines genetic material from two parent solutions to create offspring. Mutation and crossover probabilities, set as hyperparameters, determine the likelihood of these operations. The search space includes various convolutional operations ($3 \times 3$, $5 \times 5$ depth-wise separable, dilated and inverted convolutions, and $7 \times 7$ convolution), pooling operations (max and average), and skip connections. These operators, encoded within normal and reduction cells (chromosomes), alternate to form the network architecture. A normal cell extracts features with unchanged dimensions, while a reduction cell downsamples features, halving spatial dimensions to manage complexity. The encoding scheme is defined as: $\text{chromosome}(x) = LA_{1}LA_{2},LB_{1}LB_{2},LC_{1}LC_{2},LD_{1}LD_{2}$, where $L$ represents operators, and $A,B,C,D$ are indices determining connections. GA initialization uses Early Exit Population Initialization (EEPI) \cite{termritthikun2021eeea}, ensuring initial population parameters remain below a threshold $\beta$ (in millions). NAS executes for user-defined generations and population sizes, progressively discovering high-performing architectures.


Retaining the architecture search properties of EEEANet \cite{termritthikun2021eeea}, including EA with tournament selection, multiple objectives, and an early exit strategy, we incorporated driving data from the simulation. The focus is on optimizing the convolutional network component of the PPO algorithm while keeping the fully connected layers constant. Based on our previous experiments \cite{10.1007/978-3-031-82484-5_3}, we observed that improvements in RL reward performance become evident after approximately 20 epochs. Hence, to accelerate training, we use lower fidelity estimates and learning curve extrapolation, ranking architectures after 20 epochs instead of the typical 300 full RL training epochs. We also emphasize relative ranking to mitigate estimation bias. Additional lower fidelity adjustments include reducing input resolution to $84 \times 84 \times 3$, stacking 4 cells instead of 20, reducing blocks per cell from 5 to 4, and halving initial channels to 16, where blocks define combinations of convolutional operations, and initial channels set the first layer's output.  

\paragraph{In this paper we enhance EMNAS-RL methodology by two novel modules:} 
\begin{itemize}  
    \item \textbf{Optimized Transfer Learning (OTL):} From the second generation, a teacher-student framework transfers the policy of the best-performing network to student networks through Behavior Cloning (BC)\cite{inproceedings}, leveraging prior knowledge to accelerate training. Transfer learning (TL) in EMNAS aims to improve network performance across successive generations by leveraging knowledge from earlier iterations, addressing challenges unique to RL, such as the absence of labeled data and complex reward structures. Unlike supervised learning, where labeled data enables direct optimization, RL benefits significantly from TL to reduce computational costs and accelerate convergence. A two-step strategy is employed: first, networks in the current generation are pre-trained using BC with expert data from the best network of the previous generation; second, they are fine-tuned with PPO to enhance adaptability and performance. This iterative approach, starting from the second generation, ensures efficient learning throughout the evolutionary process, as shown in Figure \ref{fig:OTL}.  

    \begin{figure}[!ht]
    \centering
        \includegraphics[width = 0.55\linewidth]{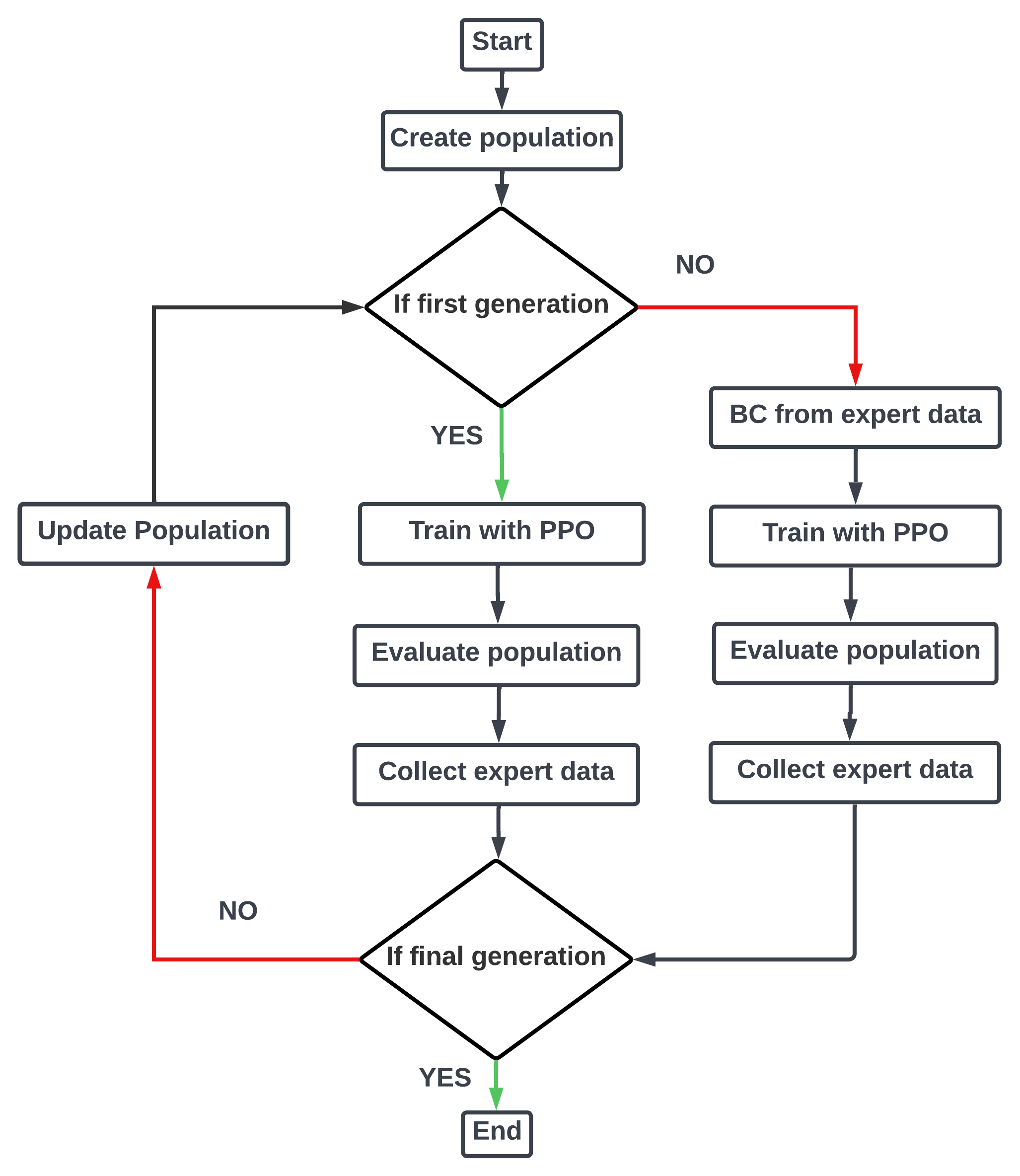}
     \caption{A flowchart illustrating the primary logic of TL, showcasing its integration as an enhancement to the EMNAS algorithm.}
    \label{fig:OTL}
    \end{figure}

    The term "Optimized Transfer Learning (OTL)" reflects the strategic process where the best-performing network transfers its learned behavior to all networks in the subsequent generation, including those retained through the survival mechanism (itself included, as it survives the evolution). This approach ensures that the BC error is uniformly propagated across the entire population, preventing any network from gaining an unfair advantage by exclusively retaining the best weights through the survival mechanism. Additionally, hyperparameter decay mechanisms retain and adjust key parameters like learning rate, PPO clip, and entropy parameters across generations, stabilizing the learning process and enabling smoother, more efficient fine-tuning. These decayed parameters are transferred to the next generation after BC, refining the networks from that point. 

    A dataset of 12,000 state-action pairs is collected from the best network of each generation, acting as an expert. To save time, state-action pairs from the final PPO iteration of the winning network are pre-saved, ensuring part of the data is immediately available. These pairs are used to pre-train the next generation's networks via BC, providing an initial policy without prior simulation interaction. Data collection begins in the second generation and continues for all subsequent generations.

    \item \textbf{Parallel Training:} EMNAS automates the parallel training of population individuals using four NVIDIA V100 32 GB GPUs, running four trainings simultaneously to optimize resource utilization.  
\end{itemize}  
  



\section{Experimental Results}
In this section, we describe the experimental setup and the impact of various hyperparameters on the performance of the NAS process. To ensure consistency across experiments, hyperparameters are kept constant: mutation probability is 0.1, crossover probability is randomized between 0.5-0.9, and survival probability is 0.2. The survival mechanism retains 1 to 4 individuals per generation, depending on population size. The number of generations and population size impact the search performance and duration, with larger values improving exploration but also increasing time. In this study, small to mid-range values are used to balance search quality and time efficiency. Additionally, the threshold parameter ($\beta$) for EEPI is varied to assess its impact. Table \ref{tab:th_search_results} shows the effect of $\beta$ on performance, with each run repeated twice for better generalization. For fixed generation (Gen) and population (Pop) values, increasing $\beta$ led to better rewards, improved parameters, and lower FLOPs. The last column shows the GPU days required to complete the NAS experiment on an NVIDIA Tesla V100-32GB, with minimal difference in resource usage despite higher values of $\beta$. Based on these results, a threshold $\beta$ of $5$ is chosen for subsequent experiments.

\begin{table}[h!]
\centering
\caption{Experiment details to compare the impact of threshold parameter ($\beta$).}
\label{tab:th_search_results}
\begin{adjustbox}{max width=0.8\textwidth}
\begin{tabular}{cccccccc}
\toprule
\textbf{Exp} & \textbf{Gen} & \textbf{Pop} &  $\boldsymbol{\beta}$ & \textbf{Reward} & \textbf{Param(M)} & \textbf{FLOPS(G)} & \textbf{GPU}\\
\midrule
1.1 & 6 & 6    & 3         & 452          & 1.13          & 1.13             &      1.0     \\
1.2 & 6 & 6      & 5    & \textbf{482}          & 1.02          & 1.01             &     1.4      \\
\midrule
2.1 & 10 & 4    & 3         & 412          & 0.99            & 0.99             &  1.2           \\
2.2 & 10 & 4  & 5       & \textbf{622}          & 0.98            & 0.93             &  1.5           \\
\bottomrule
\end{tabular}
\end{adjustbox}
\end{table}

\begin{figure}
\begin{subfigure}[!ht]{0.5\linewidth}
\includegraphics[width=\linewidth, height = 105 pt]{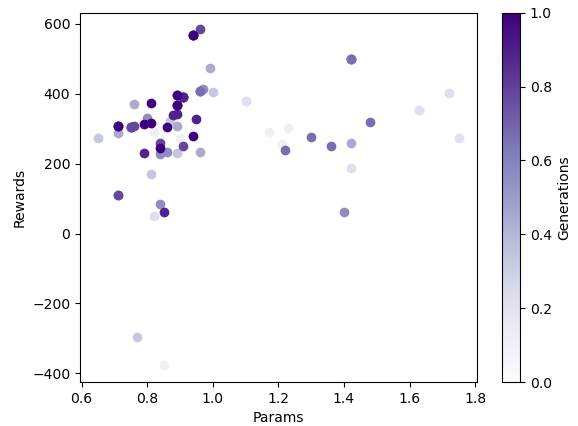}
\caption{Generation 10, Population 10}
\end{subfigure}
\begin{subfigure}[!ht]{0.5\linewidth}
\includegraphics[width=\linewidth, height = 105 pt]{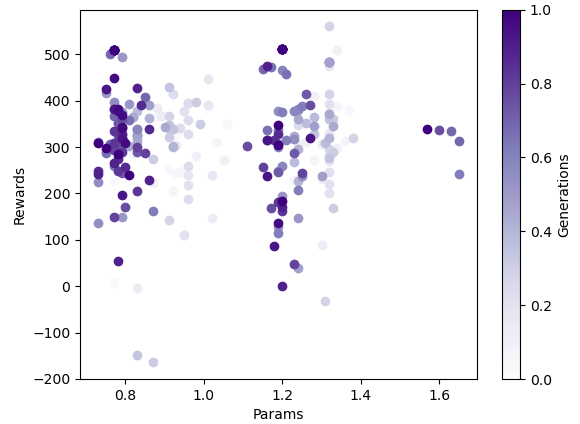}
\caption{Generation 30, Population 10}
\end{subfigure}
\caption{Evolution of rewards and number of model parameters during the architecture search over different generations. Here, the generation bar is normalized.}
\label{fig:plots}
\end{figure}

Figure \ref{fig:plots} depicts the evolution of rewards and parameters for EMNAS, color-coded by generation. To improve clarity, a 2D plot displaying only rewards and parameters is presented, with FLOPS omitted due to the similar trends observed between FLOPS and parameters. This figure indicates successful evolution, with rewards maximized and parameters and FLOPS minimized over time. Improved models consistently emerge in later generations across all experiments. These findings demonstrate the effectiveness of MOEA for NAS in identifying optimal architectures for AD. Despite variations in experiment size, no consistent advantage is observed between larger generations or population sizes, highlighting the stochastic nature of the search process. While larger experiments tend to yield better models on average across all objectives, exceptional models can still emerge from smaller searches. The increased exploration enabled by larger experiments is evident in the distinct clusters observed in Figure \ref{fig:plots}(b).

\begin{figure}[!ht]
\centering
    \includegraphics[width = 0.9\linewidth]{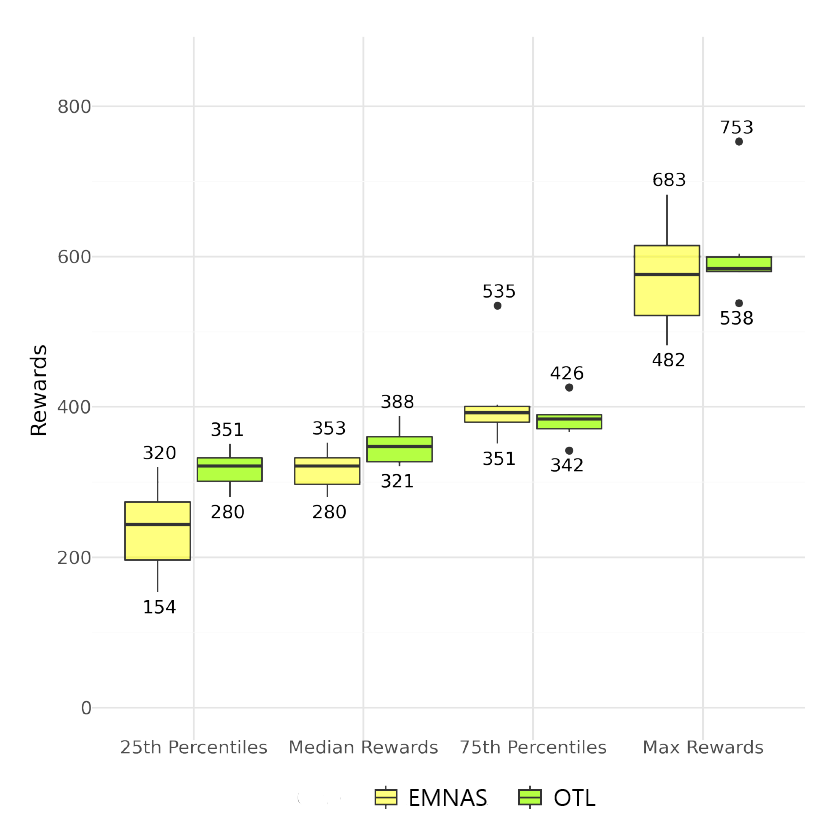}
 \caption{ Reward results presented as box plots for EMNAS (yellow) and OTL (lime), highlighting their performance distributions across various metrics.}
\label{fig:box_plot}
\end{figure}

Although EMNAS yielded satisfactory results, each generation learns from scratch. To address this, we experiment with applying the OTL teacher-student method to transfer the policy of the winning network to the next generation and evaluate whether the model performs better. Table \ref{tab:all_search_results} compares EMNAS and EMNAS with Optimized Transfer Learning (OTL) across six experimental setups, using evaluation metrics such as the 25th percentile, median, 75th percentile, and highest reward. The "P" under the method column indicates that parallelization was employed during the training process. The results clearly show that parallelization reduces runtime (in days (d)) by a factor of 2-3, with greater speed improvements observed in larger training setups. Figure \ref{fig:box_plot} provides a visual representation of the reward distributions detailed in Table \ref{tab:all_search_results}, using box plots to compare the performance of OTL and EMNAS.
\begin{table}[h!]
\centering
\caption{Reward Comparison for 6 Experiments}
\label{tab:all_search_results}
\begin{adjustbox}{max width=\textwidth}
\begin{tabular}{cccccccc}
\toprule
\textbf{Experiment} & \textbf{25th Percentile} & \textbf{Median Reward} & \textbf{75th Percentile} & \textbf{Max Reward} & \textbf{Method} & \textbf{Run Time(d)}\\
\midrule
4 populations over 10 generations  & 320 & 353 & \textbf{535} & \textbf{622} & EMNAS & 1.5 \\
2 survivors per generation         & \textbf{335} & \textbf{362} & 384 & 585 & OTL(P) & 0.7 \\
\midrule
6 populations over 6 generations   & 154 & 280 & 376 & 482 & EMNAS & 1.4\\
3 survivors per generation         & \textbf{318} & \textbf{340} & \textbf{385} & \textbf{538} & OTL(P) & 0.8  \\
\midrule
10 populations over 30 generations & 269 & 328 & 390 & 560 & EMNAS & 15.9 \\
3 survivors per generation         & \textbf{351} & \textbf{388} & \textbf{426} & \textbf{579} & OTL(P) & 6.8 \\
\midrule
15 populations over 10 generations  & 275 & \textbf{334} & \textbf{403} & 593 & EMNAS & 7.7 \\
4 survivors per generation          & \textbf{296} & 321 & 342 & \textbf{753} & OTL(P) & 3 \\
\midrule
15 populations over 15 generations & 219 & 315 & \textbf{395} & \textbf{683} & EMNAS & 13.4 \\
4 survivors per generation         & \textbf{325} & \textbf{355} & 391 & 584 & OTL(P) & 4.4 \\
\midrule
20 populations over 15 generations & 189 & 291 & 351 & 509 & EMNAS & 18.6\\
3 survivors per generation         & \textbf{280} & \textbf{323} & \textbf{367} & \textbf{604} & OTL(P) & 6.3 \\
\bottomrule
\end{tabular}
\end{adjustbox}
\end{table}

As shown in the Table \ref{tab:all_search_results} and illustrated in the box plots (Figure \ref{fig:box_plot}), OTL generally outperforms the basic EMNAS method 60\% of the time when comparing maximum rewards.  More notably, OTL consistently achieves higher median rewards, indicating greater stability compared to EMNAS, which, while capable of reaching high rewards, exhibits higher variability.  The broader interquartile range (IQR) of EMNAS highlights this variability, whereas OTL demonstrates tighter clustering of data points around the median, reflecting more consistent and reliable performance. By leveraging knowledge transfer, OTL stabilizes network performance across generations. Since rewards are the primary focus, FLOPs and parameter counts are not included in the table, though they exhibit a similar performance trend.

\begin{figure}
\begin{subfigure}[!h]{0.48\linewidth}
\includegraphics[width=\linewidth, height = 65pt]{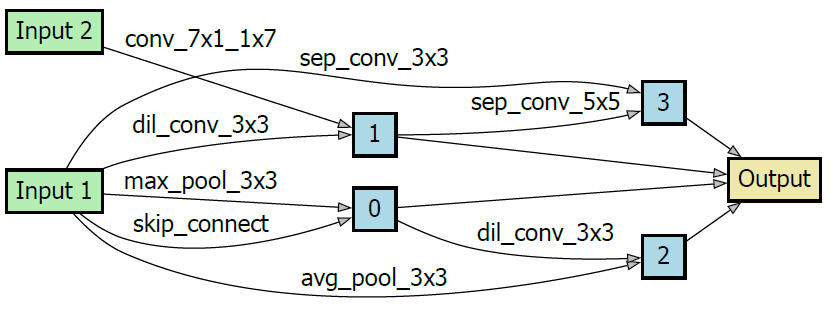}
\end{subfigure}
\begin{subfigure}[!h]{0.48\linewidth}
\includegraphics[width=\linewidth, height = 65pt]{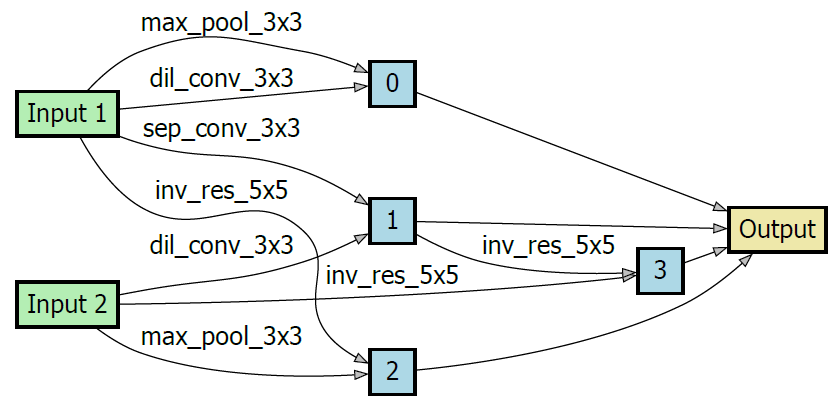}
\end{subfigure}
\caption{Normal and reduction cell structure}
\label{fig:cell_structure}
\end{figure}

Lastly, as seen in Table \ref{tab:all_search_results}, the best-performing model was found in the OTL method with a population of 15 over 10 generations, achieving a maximum reward of 753. We fully retrained the winning model over 300 PPO iterations on simulation data to assess how the architecture performs over full training. The model achieved a peak total cumulative reward of 1190, a 4\% increase over the manually set architecture (reward: 1140). Figure \ref{fig:cell_structure} illustrates the normal and reduction cell structures identified in the best-performing result of this experiment.

\section{Conclusion}
This study marks the first application of EMNAS, a multi-objective NAS framework, to large-scale RL in AD. Building on EMNAS's foundation of automating network design with evolutionary algorithms, we tailored the framework for large-scale RL and introduced parallelization to accelerate the search process by a significant factor. These advancements enabled EMNAS to consistently outperform manually designed architectures while reducing parameter counts. Additionally, to avoid learning each architecture from scratch, the OTL teacher-student method was employed to transfer policies from one generation to the next, improving stability and efficiency with significant gains in both median and maximum rewards. Further improvements are anticipated through concurrent hyperparameter and NAS optimization. Future work will focus on refining these aspects to further enhance performance.




\begin{footnotesize}


\bibliographystyle{unsrt}
\bibliography{Literatur}

\end{footnotesize}


\end{document}